\documentclass[letterpaper, 10pt, conference]{ieeeconf}      

\IEEEoverridecommandlockouts                              
\usepackage{amsmath}    
\usepackage{mathtools}    
\usepackage{amsfonts}
\usepackage{graphicx}   
\usepackage{subcaption}
\usepackage{epsfig} 
\usepackage{dsfont}
\usepackage{color}
\usepackage[normalem]{ulem}
\usepackage{cancel}
\usepackage{amssymb}
\usepackage{color}
\usepackage{bbm}
\usepackage{lipsum}
\let\labelindent\relax
\usepackage{enumitem}
\usepackage{siunitx}
\usepackage{placeins}
\usepackage{bbm}
\usepackage{float}
\usepackage{tikz}
\usetikzlibrary{positioning}
\tikzset{main node/.style={circle,fill=blue!20,draw,minimum size=1cm,inner sep=0pt},}


\usepackage[ruled,vlined,titlenotnumbered]{algorithm2e} 
\usepackage[textsize=tiny]{todonotes}

\newcommand{\R}{\mathbb{R}} 
\newcommand{\veh}{Q} 
\newcommand{\cset}{\mathcal{U}}

\newcommand{\cfset}{\mathbb{U}}

\newcommand{\reachset}{\mathcal{V}}

\newcommand{\dz}{\mathcal{Z}} 

\newcommand{\set}[1]{\{#1\}}            
\newcommand{\norm}[1]{\left\lVert#1\right\rVert}  
\newcommand{\abs}[1]{\left|#1\right|}  

\newcommand{\numsample}{L}
\newcommand{\numvehicle}{N}

\newcommand{\algo}{\mathcal{A}}
\newcommand{\numdata}{M}

\newcommand{\targetdim}{n_g}

\newcommand{\targetloc}{g}
\newcommand{\numblock}{K}
\newcommand{\samplecand}{p}
\newcommand{\numfixed}{N_{fixed}}
\newcommand{\numtrials}{N_{runs}}

\newcommand{\rom}[1]{\uppercase\expandafter{\romannumeral #1\relax}}
\usepackage{booktabs}
\usepackage{multirow}

\usepackage{varioref}
\labelformat{algocf}{\textit{alg.}\,(#1)}

\title{\LARGE \bf
Learning-based Initialization Strategy for Safety of Multi-Vehicle Systems 
}
\author{Jennifer C. Shih, Akshara Rai, Laurent El Ghaoui
    \thanks{Jennifer Shih and Laurent El Ghaoui are with the Department of Electrical Engineering and Computer Sciences, University of
     California, Berkeley. \{cshih, elghaoui\}@berkeley.edu. Akshara Rai is with Facebook AI Research. akshararai@fb.com}
}

\begin{document}
\maketitle
\thispagestyle{empty}
\pagestyle{empty}

\begin{abstract}
Multi-vehicle collision avoidance is a highly crucial problem due to the soaring interests of introducing autonomous vehicles into the real world in recent years. The safety of these vehicles while they complete their objectives is of paramount importance. Hamilton-Jacobi (HJ) reachability is a promising tool for guaranteeing safety for low-dimensional systems. However, due to its exponential complexity in computation time, no reachability-based methods have been able to guarantee safety for more than three vehicles successfully in unstructured scenarios. For systems with four or more vehicles, we can only empirically validate their safety performance. While reachability-based safety methods enjoy a flexible \textit{least-restrictive} control strategy, it is challenging to reason about long-horizon trajectories online because safety at any given state is determined by looking up its safety value in a pre-computed table that does not exhibit favorable properties that continuous functions have. This motivates the problem of improving the safety performance of \textit{unstructured} multi-vehicle systems when safety cannot be guaranteed given \textit{any} least-restrictive safety-aware collision avoidance algorithm while avoiding online trajectory optimization. In this paper, we propose a novel approach using supervised learning to enhance the safety of vehicles by proposing new initial states in very \textit{close} neighborhood of the original initial states of vehicles. Our experiments demonstrate the effectiveness of our proposed approach and show that vehicles are able to get to their goals with better safety performance with our approach compared to a baseline approach in wide-ranging scenarios.
\end{abstract}

\section{Introduction}
The safety of multi-vehicle systems has emerged as an essential and important problem as new technologies such as unmanned aerial vehicles (UAVs) develop quickly. We have seen vast interests and growth in the domain of UAVs in industry or for government purposes. For example, Google X \cite{Google2020}, Amazon \cite{Amazon20}, and UPS \cite{UPS2020} aim to use drones to accomplish their business goals of delivery of goods. Drones have also been proposed for use in transport of critical medical supplies \cite{Zipline2020}, \cite{Vayu2020}. There have also been many efforts in using UAVs for disaster responses and military operations \cite{DSLRPros2020}, \cite{Humanitarion2020}, \cite{AUVSI16}, \cite{Military2020}. Due to the substantial growth in utilizing drones for a wide range of domains, the Federal Aviation Administration created guidelines specifically targeting UAVs in recent years \cite{FAA2020}. It is thus of high urgency to develop effective approaches for multiple UAVs to achieve their goals in the same environment safely. A central problem in this realm is the ability to have robots visit their goals safely starting from some approximate initial regions. The ability to pick exact initial conditions that makes the multi-vehicle system safer from an approximate initial regions for the vehicles is thus highly crucial.

The problem of safety in multi-agent systems has been studied through various approaches. Some methods used potential functions to address safety while vehicles travel along pre-specified trajectories \cite{Saber02, Chuang07}. There have also been works that introduce the idea of velocity obstacles, induced by control inputs of the vehicles, for collision avoidance \cite{Mora15, Fiorini98, Vandenberg08}. Authors in \cite{Mastellone08} used control strategies derived with Lyapunov-type analysis for safe control of multiple vehicles. However, these approaches do not flexibly offer the safety guarantee for general dynamical systems that reachability offers. They also do not offer the desirable property of a ``least-restrictive" safe control strategy that reachability-based strategies permit.

A promising class of methods for addressing safety in the context of multi-vehicle systems is differential games. In particular, Hamilton-Jacobi (HJ) reachability \cite{Mitchell05} is a framework that uses differential games to model conflicts of more than one agent. However, although HJ reachability offers safety guarantees for general dynamical systems, its computation scales exponentially with the number of states in the systems, limiting reachabilty to be directly applicable to systems with only two vehicles \cite{Fisac15, Mitchell05}. While attempts have been made to use reachability-based methods to guarantee safety for a larger number of vehicles \cite{Chen15, Chen15b, Shih2021}, these works either make strong assumptions on the formation of the vehicles or require that the vehicles know other vehicles' trajectories \textit{a priori}. In contrast, in this paper, we tackle \textit{unstructured} collision avoidance where \textit{unstructuredness} refers to the scenario that vehicles do not have to follow specific structures and formation or require knowledge of future trajectories of other agents. 

The recent work \cite{Chen16} is the first work that enables guaranteed safety for three vehicles in unstructured settings using reachability via a higher level control logic. \cite{Chen17} further investigates the problem of guaranteed safety for four vehicles, however, it requires the assumption that vehicles can be removed from the environment when conflicts cannot be resolved, which is not always possible. Guaranteed-safe collision avoidance methods for four or more vehicles in unstructured settings without needing to remove vehicles in certain situation using reachability do not yet exist. However, reachability-based methods enable the desirable least-restrictive safe control algorithms such that agents can perform \textit{any} action while they're deemed safe. Inspired by this, we tackle the problem of improving safety performance of systems with at least four vehicles when the vehicles adopt least-restrictive safe control strategies. 
While our goal is not to offer safety guarantees, we demonstrate that our proposed learning-based approach can effectively improve safety performance just by learning good initialization of the vehicle states.

Machine learning approaches for tackling collision avoidance for multi-vehicle systems have been investigated in prior works. For example, \cite{long2017} uses an end-to-end learning approach to generate reactive safe policies. However, it only considers local collision avoidance and assumes the system is holonomic. Another line of work uses reinforcement learning (RL) to learn control policies of multi-vehicle systems \cite{Busoniu08}, \cite{Gupta2017}. However, RL-based methods require substantial number of experiences of interactions among the vehicles to learn good policies and can take hours and, often, days to train. Furthermore, they do not result in least-restrictive safe controllers. In contrast, our proposed method is not aim at learning a policy but directed towards tackling the problem of improving safety performance through learning better initialization. 

Our main contribution is a novel learning-based approach to effectively enhance the safety of multi-vehicle systems by learning good initialization of vehicles. We formulate the problem such that each vehicle is tasked with visiting a goal and each also proposes a state it will start closely at. These agents use a least-restrictive safety-aware algorithm to get to their goals while taking safety into account. Motivated by the fact that safety cannot be guaranteed for larger multi-vehicle systems and the difficulty of reasoning about long-horizon trajectories for least-restrictive safety-aware algorithms, we show that it is possible to figure out, without human intervention, a fast and effective strategy that makes only \textit{minor} modification to each agent's original proposed initial state and run the same safety-aware algorithm while improving the safety performance of the system. We demonstrate through extensive experiments on four to six vehicles that our proposed learning-based method consistently and reliably improves the safety performance of multi-vehicle systems. 

\section{Background}
\label{sec:background}
In this paper, we propose to use machine learning to learn initialization of multi-vehicle systems to effectively enhance the safety of the systems when vehicles adopt a least-restrictive safety-aware algorithm to get toward their goals. In this section, we provide an overview of HJ reachability to motivate what is means for a safe control strategy to be least-restrictive, a discussion on least-restrictive safe control methods, and the relevant machine learning background for our proposed approach.

\subsection{Hamilton-Jacobi (HJ) Reachability \label{sec:HJI}}

HJ reachability is a promising method for ensuring safety. We give a brief overview of how HJ reachability is used to guarantee safety for a pair of vehicles as presented in \cite{Mitchell05}. For any two vehicles $\veh_i$ and $\veh_j$ with dynamics describe by the following ordinary differential equation (ODE)
\begin{equation}
\label{eq:vdyn} 
\dot{x}_m = f(x_m, u_m), \quad u_m \in \cset, m = i, j, \\
\end{equation} 
their relative dynamics can be specified by an ODE
\vspace{-0.5em}
\begin{equation}
\label{eq:rdyn} 
\begin{aligned}
\dot{x}_{ij} &= g_{ij}(x_{ij}, u_i, u_j), u_i,u_j \in \cset
\end{aligned}
\end{equation}
where $x_{ij}$ is a relative state representation between $x_i$ and $x_j$.

In the reachability problem, for any pair of vehicles $\veh_i$ and $\veh_j$, we are interested in determining the backward reachable set (BRS) $\reachset_{ij}(T)$, the set of states from which there exists no control for $\veh_i$, in the worst case non-anticipative control strategy by $\veh_j$, that can keep the system from entering some final set $\dz_{ij}$ within a time horizon $T$. For safety purpose, $\dz_{ij}$ represents dangerous configurations between $\veh_i$ and $\veh_j$. In this paper, we assume the danger zones $\dz_{ij}$'s are defined such that $x_{ij} \in \dz_{ij} \Leftrightarrow x_{ji} \in \dz_{ji}$. Note in this paper we will also use the term unsafe set to refer to the backward reachable set.

The backward reachable set $\reachset_{ij}$ can be mathematically described as
\begin{equation}
\label{eq:brs}
\begin{aligned}
&\reachset_{ij}(t) = \{x_{ij}: \forall u_i \in \cfset, \exists u_j \in \cfset, \\
&x_{ij}(\cdot) \text{ satisfies \eqref{eq:rdyn}}, \exists s \in [0, t], x_{ij}(s) \in \dz_{ij}\},
\end{aligned}
\end{equation}
and obtained by $\reachset_{ij}(t) = \{x_{ij}: V_{ij}(t, x_{ij}) \le 0\}$ where the details on how to obtain the value function $V_{ij}(t, x_{ij})$ is in \cite{Mitchell05}. In this paper, we assume $t \rightarrow \infty$ and write $V_{ij}(x_{ij}) = \lim_{t \rightarrow \infty} V_{ij}(t, x_{ij})$. If the relative state $x_{ij}$ of $\veh_i$ and $\veh_j$ is outside of $\reachset_{ij}$, then $\veh_i$ is safe from $\veh_j$. If $x_{ij}$ is at the boundary of $\reachset_{ij}$, \cite{Mitchell05} shows that as long as the optimal control
\begin{equation}
    u_{ij}^* = \arg \max_{u_i \in\cset} \min_{u_j \in\cset} D_{x_{ij}} V(x_{ij}) \cdot g_{ij}(x_{ij},u_i,u_j)
    \label{eq:opt_action}
\end{equation}
is applied immediately, $\veh_i$ is guaranteed to be able to avoid collision with $\veh_j$ over an infinite time horizon.

\subsection{Least-restrictive safe control strategies}
As we've seen in Section \ref{sec:HJI} on HJ reachability, as long as the optimal safe control in Equation \eqref{eq:opt_action} is applied by $\veh_i$ at the boundary of the BRS  $\reachset_{ij}$, $\veh_i$ will remain safe from $\veh_j$ for all time. A similar least-restrictive safe control strategy based on reachability can be adopted for single agent systems that aim to avoid dangerous regions in the environment while disturbance is present. This enables a least-restrictive control strategy where an agent gets to execute any type of controller such as a goal controller that gets the vehicle to its target \cite{Chen16}, \cite{Chen17} or a machine learning-based controller  \cite{Akametalu2014}, \cite{Shih2020} when the agent is not at the boundary of a backward reachable set.

This can be a highly desirable property in a safety strategy because it offers high flexibility for agents to execute whatever control they would like when they're deemed safe and decouples the reasoning of safe controllers and task-oriented controllers. On the other hand, this means that the system performs a zero-step look-ahead at each time step online and does not reason how its current action affect the future trajectories of all the vehicles in the environment. Theoretically we can incorporate safety derived from reachability into a trajectory optimization problem and reason about future trajectories online. However, it would be very difficult to perform this optimization online efficiently due to the nature that the value function $V_{ij}(x_{ij})$ is a discrete look-up table computed offline and hence does not exhibit favorable properties that continuous functions have in terms of being easily incorporated into an optimization problem and optimized efficiently. One could theoretically perform the optimization in an optimization problem that requires look-up of values in a discrete table with a sampling-based method, however, this scales poorly with the time horizon of the trajectory and the number of vehicles. 

Motivated by the difficulty of planning and reasoning about long horizon trajectories online with least-restrictive safe control strategies such as those based on reachability, we instead investigate whether it is possible to learn a good initialization strategy that identifies a new set of initial states within a \textit{close} neighborhood of some original proposed initial states for all agents in the environment so that the safety performance of the multi-vehicle system improves without having to reason explicitly about future trajectories. Interestingly, we demonstrate in this paper that this is possible. 

\subsection{Probabilistic Modeling in Machine Learning}
\label{sec:background:ml}
Machine learning (ML) has emerged as a promising tool for many application domains such as vision, speech, and robotics. ML methods have shown high potential in tackling problems when the dimension of the task is high or when direct modeling is not feasible due to the complexity of the system and the substantial computation required. 

Given a data set $\set{(h_j, y_j)}_{j=1}^{n}$ where $h_j$ is the feature vector and $y_j$ is the corresponding label for each data point, a classification problem aims to learn how to best predict the label given the feature vector through training. In this paper, we focus on the scenario where the labels $y$ are binary labels that take on values of either $0$ or $1$. We adopt ML models that are parameterized by some function $f_{\theta}(h)$ where $\theta$ is the parameter we aim to learn. In particular, we model the probability that feature $h$ has label $y=1$ as $f_{\theta}(h)$, i.e., $p(y=1|h) = f_{\theta}(h)$ and $p(y=0|h) = 1 - f_{\theta}(h)$. The function approximator $f_{\theta}(h)$ can in general be any model such as a neural network.

By assuming each observation is independent and identically distributed, a common assumption in probabilistically-motivated ML loss functions, we have that the likelihood of observing the data points $\set{(h_j, y_j)}_{h=1}^{n}$ given $\theta$ is
\begin{equation}
    L(\theta) = \prod_{j=1}^{n} f_{\theta}(h_j)^{y_j} (1 - f_{\theta}(h_j))^{1- y_j}.
\end{equation}

The goal of learning is to then maximize the above likelihood with respect to $\theta$. This is equivalent to minimizing the negative log likelihood on the data set with respect to $\theta$: 
\begin{equation}
    \sum_{j=1}^n - y_j \log f_{\theta}(h_j) - (1-y_j) \log \left(1 - f_{\theta}(h_j)\right).
\end{equation}

\section{Problem Formulation \label{sec:formulation}}
Consider $N$ vehicles, denoted $\veh_i, i = 1, 2, \ldots, N$, with identical dynamics described by the following ordinary differential equation (ODE)
\begin{equation}
\label{eq:vdyn} 
\dot{x}_i = f(x_i, u_i), \quad u_i \in \cset, \quad i = 1,\ldots, N
\end{equation}
\noindent where $x_i \in \R^{n}$ is the state of the $i$th vehicle $\veh_i$, and $u_i$ is the control of $\veh_i$. Each of the $N$ vehicles is tasked with visiting a target whose location $\targetloc_i \in \mathbb{R}^{\targetdim}$ is known before all vehicles begin their journey. 

We assume the vehicles adopt a least-restrictive safety-aware algorithm $\algo$ that explicitly optimizes for safety of the vehicles while they get to their targets. Given the vehicle dynamics in Equation \eqref{eq:vdyn}, the initial states $x_i(t_0)$, and the target location $\targetloc_i$ of each vehicle $\veh_i$, the algorithm $\algo$ should determine the control $u_i$ for each vehicle $\veh_i$ based on the joint configuration of all vehicles at each time step. In addition, the safety-aware algorithm $\algo$ should be primarily designed with safe control of multiple vehicles in mind and should not be naive when concerning safety. For example, the algorithm introduced in \cite{Chen16} satisfies this criteria. In this paper, we focus on multi-vehicle systems where there is no guarantee that the safety-aware algorithm $\algo$ is able to get all vehicles to their targets without any safety violation and our goal is to improve the safety performance of the system when safety cannot be guaranteed.

We allow each vehicle the flexibility in determining approximately where their starting states are. Instead of having full freedom of placing vehicles wherever we wish, we make the problem more challenging by only allowing our proposed method to place each vehicle within a \textit{close} neighborhood of the original proposed state of the vehicle. We also enact the constraint that we are not allowed to modify the proposed initial states of $\numfixed$ vehicles of the $\numvehicle$ vehicles. Mathematically, let the original proposed initial state of vehicle $\veh_i$ be $x_{i,o}(t_0) = [p_{1,o}, p_{2,o}, \dots, p_{\numblock,o}]$ where $p_{k,o}$'s, $k \in \set{1, \dots, \numblock}$, are disjoint blocks of the state and how the state of the system is divided into blocks can be freely determined by users of our proposed approach.
For each agent $\veh_i$ such that we are allowed to modify the initial state for, the new initial state $x_{i}(t_0) = [p_{1}, p_{2}, \dots, p_{\numblock}]$ based on our proposed method should satisfy constraints $\norm{p_{k} - p_{k,o}} \leq \epsilon_k$ for some small real $\epsilon_k > 0$. The norm can be any norm that makes sense for measuring distance, which typically we use the L-1 or L-2 norm. On the other hand, if we are restricted from modifying a vehicle $\veh_j$'s initial state, then the new initial state $x_{j}(t_0) = x_{j,o}(t_0)$.

Given the vehicle dynamics in Equation \eqref{eq:vdyn}, the original proposed initial state of each vehicle $x_{i,o}(t_0)$, the set of $\numfixed$ vehicles that we cannot modify initial states for, the danger zones $\dz_{ij}$, the target location $\targetloc_i$ for each vehicle $\veh_i$, and the least-restrictive safety-aware algorithm $\algo$, we propose an effective learning-based method to improve the safety performance of the multi-vehicle system while adopting the same algorithm $\algo$. 

We demonstrate the effectiveness of our proposed learning-based approach by comparing it with randomly selecting \textit{close} neighboring states of the original proposed initial states as new initial states with experiments and show that our approach results in better overall success rate of zero safety violation throughout the execution. Our proposed method also achieves lower number of total safety violations on average.  

\section{Methodology \label{sec:method}}
In this section, we describe in detail our proposed learning-based method for improving safety performance of \textit{any} least-restrictive safety-aware algorithm while incurring very little computation cost online. In particular, our proposed approach encompasses how we frame this problem as a machine learning problem, which includes gathering data, modeling the problem, learning the model, and using the learned model to obtain better initialization for the vehicles to enhance the safety performance of the multi-vehicle system. 

\subsection{Data gathering and preparation \label{subsec:data}}
To gather training data for a $\numvehicle$-vehicle system for our proposed approach, we ran $\numdata$ simulations such that the initial states of all vehicles are randomly generated as follows. First, we determine $\numvehicle$ distinct initial states that will likely make collision avoidance a challenging problem. In each simulation, we then randomly assign each vehicle to a distinct initial state it should start close to. For each vehicle, we further randomly sample a state around the initial state it is assigned to such that the new initial state is in close proximity to its original initial state as illustrated in Section \ref{sec:formulation}. There are $\numvehicle$ distinct fixed goal locations, one for each of the $\numvehicle$ vehicles. In each simulation, we also randomize the goal location each vehicle is assigned to. The reason we determine in advance a set of original initial states the vehicles should start close to and the target locations instead of just randomly sample initial states and target locations throughout the entire space is that the vehicles will rarely even come close to being in danger of each other in the latter initialization method. We want to focus on challenging scenarios where we have high confidence that the agents will come into close contact with each other as they head to their targets, enabling the safety-critical control from algorithm $\algo$ to play a large role in the safety performance and making it meaningful to apply our proposed method.

For each simulation $j \in \set{1, \dots, \numdata}$, we keep track of the following information: the initial states of all vehicles $x_i(t_0)$'s, the goal locations of all vehicles $\targetloc_i$'s, and an indicator variable $y_j$ that represents whether the least-restrictive safety-aware algorithm $\algo$ was able to get all vehicles to their goals without \textit{any} vehicle getting into each other's danger zone in this trial. We let $y_j = 1$ if all vehicles reach their goals without \textit{any} safety violation and $y_j = 0$ otherwise. Note that for each trial $j$, the simulation continues even when vehicles get into each other's danger zone and only ends when all vehicles have reached their goals.

To illustrate how we propose to construct the features for training, first let the concatenated vector of all initial states $x_i(t_0)$ and target locations $\targetloc_i$ in trial $j$ be $p_j = [x_1(t_0), \dots, x_{\numvehicle}(t_0), \targetloc_1, \dots, \targetloc_{\numvehicle}] \in \mathbb{R}^{\numvehicle \times (n+n_g)}$. We construct the feature map $\phi(p)$ as follows: first we determine the order these initial states are in counter-clockwise starting from a particular reference direction such as the twelve o'clock direction. This gives a bijective map whose domain and range are both $\set{1, \dots, N}$ and maps each vehicle to its position based on the ordering logic. We use $x^{(i)}(t_0)$ and $\targetloc^{(i)}$ to denote the initial state and the target location of the vehicle in the $i$th position based on the ordering logic mentioned above. The feature map $\phi$ is then 
\begin{equation}
    h = \phi(p) = [x^{(1)}(t_0), \dots, x^{(\numvehicle)}(t_0), \targetloc^{(1)}, \dots, \targetloc^{(\numvehicle)}].
\end{equation}
After applying this feature map to the data gathered from all $\numdata$ trials, we obtain the data set $\set{h_j, y_j}_{j=1}^{\numdata}$ for training. In general, $\numdata$ is selected based on the user of the proposed algorithm to provide sufficient data to learn a good model and is dependent on the dynamics and the size of the multi-vehicle system.

\subsection{Learning a model with machine learning \label{subsec:learning}}
To achieve our desired goal of determining good initialization for vehicles, one intermediate step is to determine the likelihood of algorithm $\algo$ succeeding in getting all vehicles to their targets without \text{any} vehicle getting into each other's danger zone given the initial states and target locations of all vehicles. To achieve this, we use supervised learning to make predictions on this likelihood.

During training, we model the probability that algorithm $\algo$ will succeed in getting all vehicles to their targets without any safety violations given the feature vector $h$ as $f_{\theta}(h)$ and aim to learn the parameter $\theta$. As described in Section \ref{sec:background:ml}, given data set $\set{h_j, y_j}_{j=1}^{\numdata}$ where $y_j$'s are binary variables, we minimize the following negative log likelihood with respect to $\theta$ to solve for the optimal $\theta$:
\begin{equation}
    \sum_{j=1}^n - y_j \log f_{\theta}(h_j) - (1-y_j) \log \left(1 - f_{\theta}(h_j)\right).
\end{equation}

We use stochastic gradient descent to find a local minimizer $\theta^{\star}$ of this loss function. Once we obtain the minimizer $\theta^{\star}$, given any new feature vector $h$ representing the configuration of the initial states and target locations of the vehicles in the environment not seen during training, we predict the probability that algorithm $\algo$ will get all vehicles to their goals without any safety violations as $f_{\theta^{\star}}(h)$.

\subsection{Evaluation on novel test scene online}
\label{method:eval}
Recall that our goal is to design a strategy to propose a new initial state \textit{close} to the original proposed initial state of each vehicle that will result in a higher success rate of getting all vehicles to their targets without any danger zone violations. We also aim to have less number of total danger zone violations with a better initialization.

Given a novel scene online where the original proposed states of each vehicle $\veh_i$ is $x_{i,o}(t_0) = [p_{1,o}, p_{2,o}, \dots, p_{\numblock,o}]$ where as described in Section \ref{sec:formulation}, $p_{k,o}$'s are disjoint blocks of the state. Suppose each vehicle's target location is $\targetloc_i$. To find a good initialization, first we uniformly sample $\numsample$ sets of $\numvehicle$ initial  states in the close neighborhood of $x_{i,o}(t_0)$'s by setting the constraint that each sampled state $x_{i}(t_0)$ should be in the set $\set{x_{i}(t_0) = [p_{1}, p_{2}, \dots, p_{\numblock}] | \norm{p_{k} - p_{k,o}}\leq \epsilon_k, k = 1, \dots, \numblock}$. For the $\numfixed$ vehicles that we are restricted from modifying their original proposed initial states of, we set $\epsilon_k = 0, \forall k \in \set{1, \dots, \numblock}$. Otherwise we set $\epsilon_k$ to a small positive real number. Note that we cannot modify the target locations of the vehicles. 

After obtaining $\numsample$ candidate sets of initial states for all vehicles and constructing data points $\samplecand_l$'s, $l \in \set{1, \dots, \numsample}$, based on Section \ref{subsec:data}, we select the set of initial states with the highest likelihood of succeeding in getting all vehicles to their targets without any danger zone violations using the learned function approximator $f_{\theta^{\star}}(h)$ as the new set of initial states. Mathematically, the selected set of initial states is the the set of initial states $\samplecand^{\star}$ correspond to where 
\begin{equation}
    \samplecand^{\star} = \underset{l \in \set{1, \dots, \numsample}}{max} \text{ } f_{\theta^{\star}}(\phi(\samplecand_l)).
\end{equation}

The number of candidate sets $\numsample$ can in general be a large number as the above computation can be easily done in parallel.

\section{Experiments}
\label{sec:simulation}
In this section, we present extensive experimental results which demonstrate that with our proposed learning-based initialization strategy, the safety performance of multi-vehicle systems are effectively and reliably better compared with the baseline initialization strategy that randomly picks a set of initial states from all candidate initial sets in the vicinity of the original proposed set of initial states of all vehicles. 

For all experiments, we use the least-restrictive safety-aware algorithm proposed in \cite{Chen16}. This reachability-based algorithm guarantees safety for three-vehicle systems but does not guarantee safety when the number of vehicles $\numvehicle$ is greater than $3$. This algorithm has been demonstrated to be substantially better in safety performance already compared to a baseline safety algorithm in the paper when $\numvehicle > 3$. Thus the safety algorithm we use in this paper is not a naive collision avoidance algorithm and is an ideal algorithm for use in the evaluation on our proposed learning-based initialization strategy. We conduct experiments using this algorithm on multi-vehicle systems where the number of vehicles $\numvehicle$ are equal to 4, 5, and 6.

In our experiments, the dynamics of each vehicle $\veh_i$ is given by the Dubins Car dynamics
\vspace{-0.5em}
\begin{equation*}
\dot{q}_{x,i} = v \cos \theta_i, \text{ } \dot{q}_{y,i} = v \sin \theta_i, \text{ } \dot{\theta}_i = \omega_i, \quad |\omega_i| \le \bar{\omega}
\vspace{-0.5em}
\end{equation*}
\noindent where the state variables $q_{x,i}, q_{y,i}, \theta_i$ represent the $x$ position, $y$ position, and heading of vehicle $\veh_i$. Each vehicle travels at a constant speed of $v=5$, and chooses its turn rate $\omega_i$, constrained by maximum $\bar{\omega}=1$. The danger zone for HJ computation between $\veh_i$ and $\veh_j$ is defined as
\begin{equation}
\dz_{ij} = \{x_{ij}: (q_{x,j} - q_{x,i})^2 + (q_{y,j} - q_{y,i})^2 \leq R_c^2\},
\end{equation}
\noindent whose interpretation is that $\veh_i$ and $\veh_j$ are considered to be in each other's danger zone if their positions are within $R_c$ of each other. Here, $x_{ij}$ represents their joint state, $x_{ij} = [q_{x,j} - q_{x,i}, q_{y,j} - q_{y,i}, \theta_j - \theta_i]$.

To evaluate the effectiveness of our proposed learning-based initialization strategy, we perform large scale experiments on two settings of the speed $v$ in the dynamics and the danger zone radius $R_c$, ($v=5, R_c=5$) and ($v=6, R_c=4$), for number of vehicles $\numvehicle = 4, 5, 6$. For each set of experiments, we evaluate extensively the safety performance of the multi-vehicle systems by varying $\numfixed$, the number of vehicles such that we cannot modify the original proposed initial states for. Note that it only makes sense to run experiments to evaluate the effective of our approach when we can modify at least one of vehicles' proposed initial states. For each run, we initialize each vehicle by placing them symmetrically on a circle of radius $10 + 2 \times (\numvehicle -3)$ facing the center of the circle, and then add random perturbations to these states in each run. This gives us the original proposed initial states of all vehicles. This initialization ensures challenging collision avoidance scenarios as all vehicles will likely come in close contact with each other as they head to their targets. For any vehicle $\veh_i$ that we are allow to modify its initial state for, suppose its proposed initial state is $x_{i,o}(t_0) = [q_{x,i,o},q_{y,i,o}, \theta_{i,o}]$. We constrain the new initial state $x_{i}(t_0) = [q_{x,i}, q_{y,i}, \theta_{i}]$ to be in close proximity to the original proposed state such that $x_{i}(t_0)$ should satisfy
\begin{equation}
    \abs{q_{x,i,o}-q_{x,i}} \leq 3, \abs{q_{y,i,o}-q_{y,i}} \leq 3, \abs{\theta_{i} - \theta_{i,o}} \leq \pi /5.
    \label{eq:constraint}
\end{equation}

We train a model for each $\numvehicle \in \set{4,5,6}$ for each of the two settings ($v=5, R_c=5$) or ($v=6, R_c=4$). To train each model, we gather $5000$ data points when $\numvehicle=4$ and $10000$ data points when $\numvehicle=5,6$ with the data collection technique proposed in Section \ref{subsec:data}. The hyperparameter settings are described as follows. For all models, we use a three-layer fully connected neural network with ReLU activation on the hidden layer and Sigmoid activation on the output layer. The number of nodes in the hidden layer of the network is $10, 15, 20$ for $\numvehicle=4,5,6$ vehicles respectively. As described in Section \ref{subsec:learning}, we use the cross entropy loss. For optimization, we the Adam optimizer \cite{kingma14} with learning rate $0.01$ to train the network.

To demonstrate the safety benefits with our proposed approach, we compare our proposed initialization strategy with a random initialization strategy. For each $\numvehicle$-vehicle system, we ran $\numtrials=200$ randomized runs for each possible combination of $v, R_c, \numfixed$. For each individual comparison run, we randomly sample $\numsample = 10$ set of initial states in close proximity to the original proposed states such that each set satisfies the constraints \eqref{eq:constraint}. With our proposed learning-based approach, we use the learned parameters of ML model to select the best set of proposed initial states out of all candidate sets; for the baseline random initialization strategy, we uniformly select one of the $\numsample$ candidate sets of initial states as the initial states for the vehicles. We report the results for the settings ($v=6, R_c=4$) and ($v=5, R_c=5$) in Table \ref{tab:speed_6_r_4} and Table \ref{tab:speed_5_r_5} respectively. We consider the following two safety metrics:
\begin{itemize}
    \item Success rate $p_s$: the percentage of runs such that \textit{all} vehicles get to their goals without \textit{any} safety violation.
    \item Average number of collisions $N_{col}$: the total number of safety violations throughout the \textit{entire} execution for all $\numtrials$ runs divided by the product of the number of vehicles $\numvehicle$ and the number of runs $\numtrials$. One safety violation is defined as a pair of vehicle being within a distance $R_c$ of each other in a time step. There can be multiple safety violations at a given time because multiple pairs of vehicles might be in each other's danger zones at once.
\end{itemize}

From Table \ref{tab:speed_6_r_4} and Table \ref{tab:speed_5_r_5}, we can see that our proposed learning-based initialization strategy effectively and reliably improves the success rate $p_s$ and reduces the average number of collisions $N_{col}$ across all experiments, For some scenarios, our approach substantially outperforms the baseline. In particular, when $v=5, R_c=5, \numvehicle=4, \numfixed=1$, we see that our proposed approach has $90\%$ success rate whereas the baseline approach has only $66.5\%$ success rate. In the same setting, the average number of collisions with our proposed approach is only $25\%$ of the average number of collisions with the baseline initialization strategy. In general, we can see that our proposed approach outperforms the baseline approach more substantially when $\numfixed$ is smaller, which intuitively makes sense because we get to optimize the states of more vehicles when $\numfixed$ is small. In addition, we can also observe that our proposed approach is more likely to considerably outperform the baseline when the number of vehicles $\numvehicle$ is smaller, which also makes intuitive sense as there are likely more interactions among vehicles that are more difficult to be inferred by the initial states alone when $\numvehicle$ is large. However, even when the number of vehicles $\numvehicle$ is $6$ and $\numfixed=5$, we still get around an $18\%$ reduction in the average number of collisions for both settings of $(v, R_c)$ with our proposed learning-based initialization method.

\begin{figure}[]
\centering
  \begin{subfigure}[b]{0.23\textwidth}
    \includegraphics[width=\textwidth]{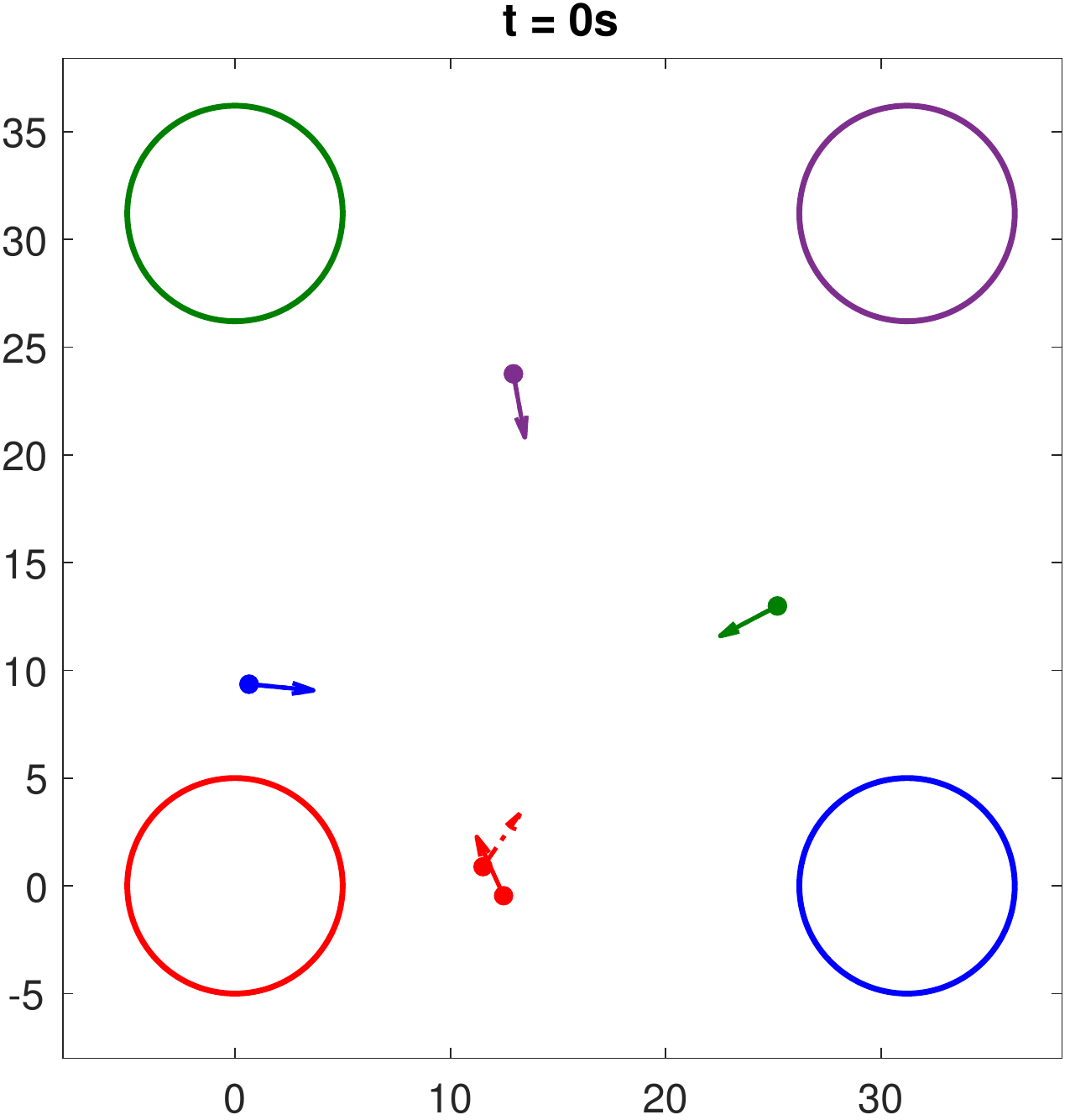}
  \end{subfigure}
  \begin{subfigure}[b]{0.23\textwidth}
    \includegraphics[width=\textwidth]{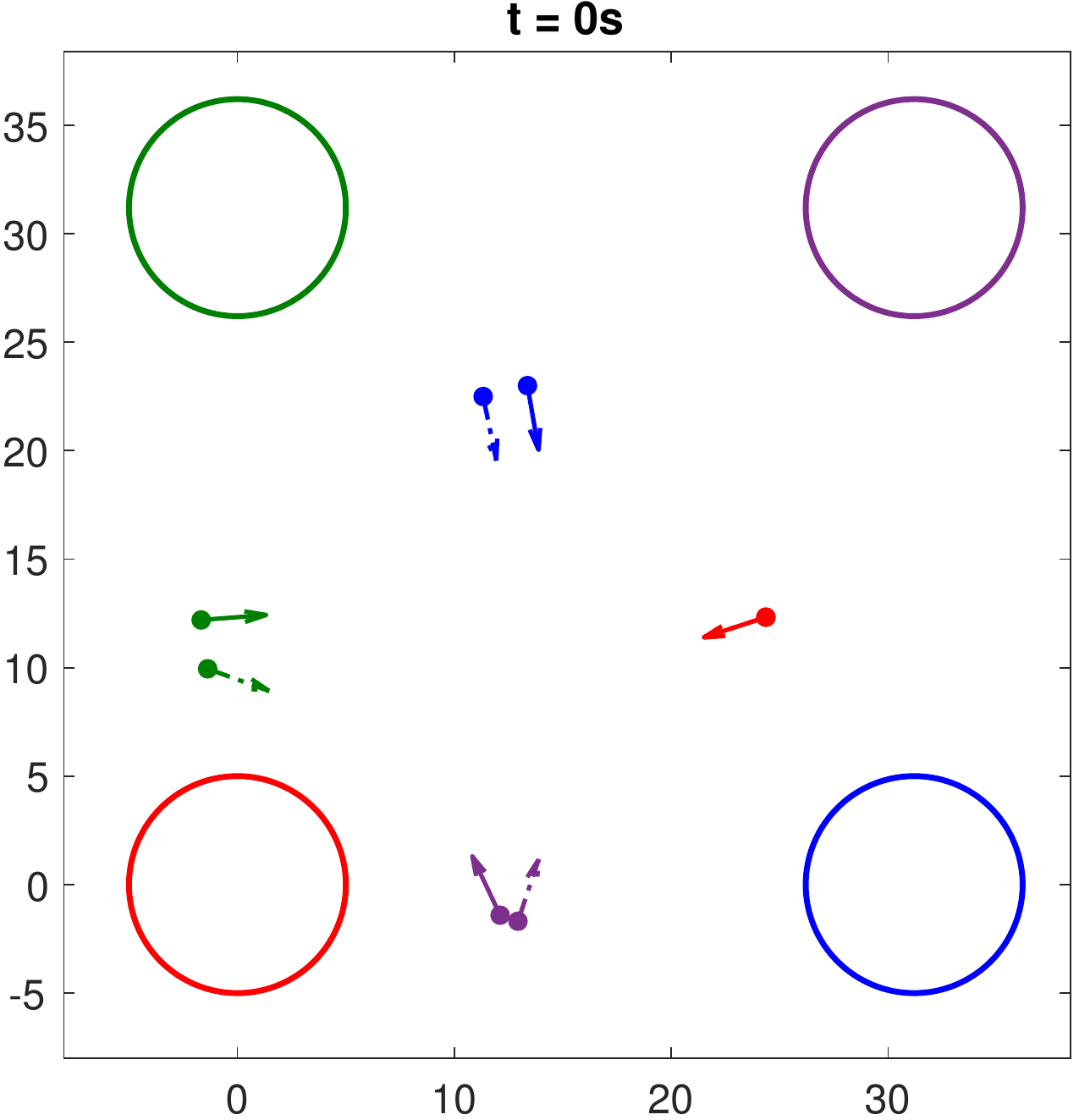}
  \end{subfigure}
  \\
  \begin{subfigure}[b]{0.23\textwidth}
    \includegraphics[width=\textwidth]{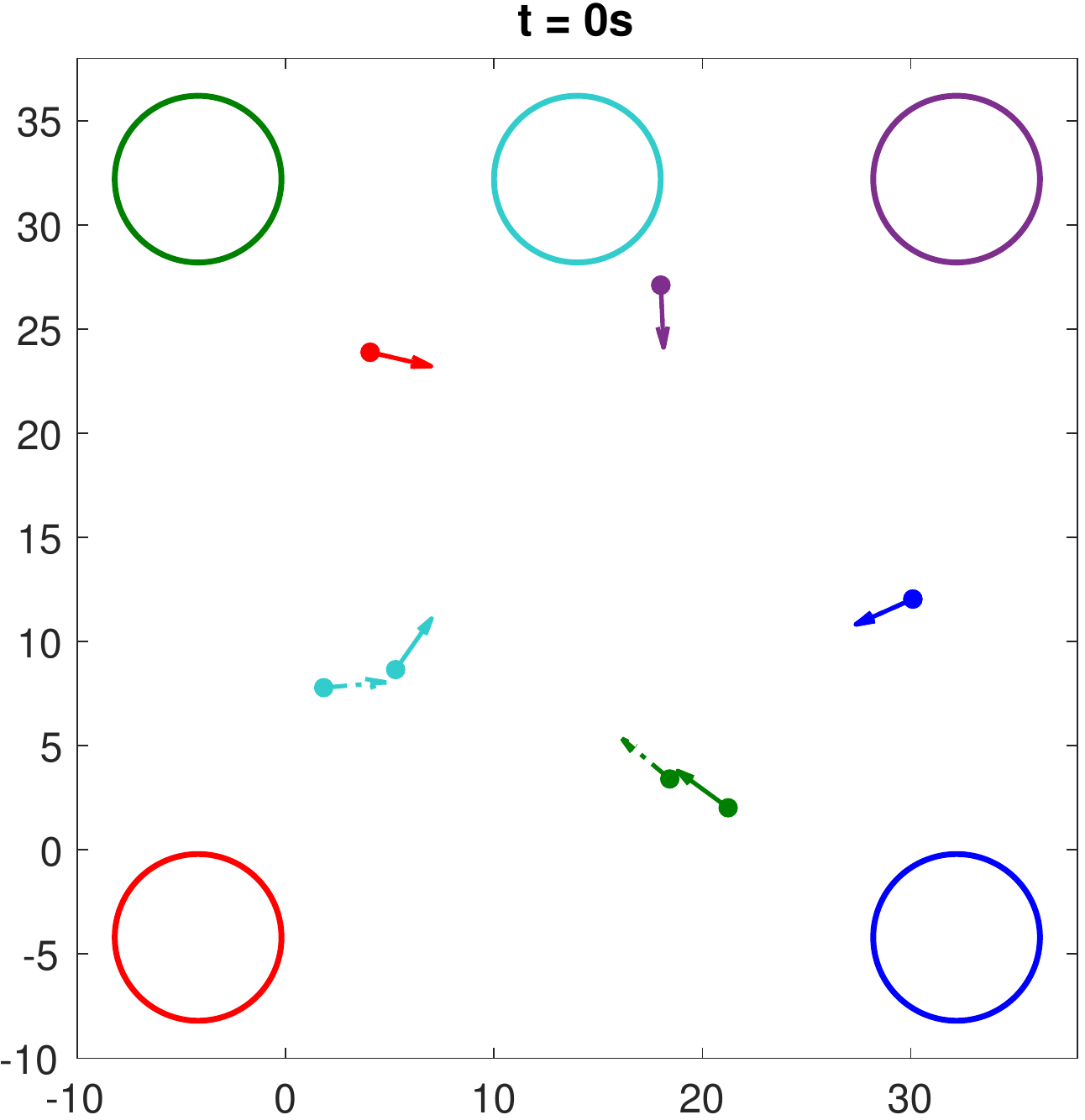}
  \end{subfigure}
  \begin{subfigure}[b]{0.23\textwidth}
    \includegraphics[width=\textwidth]{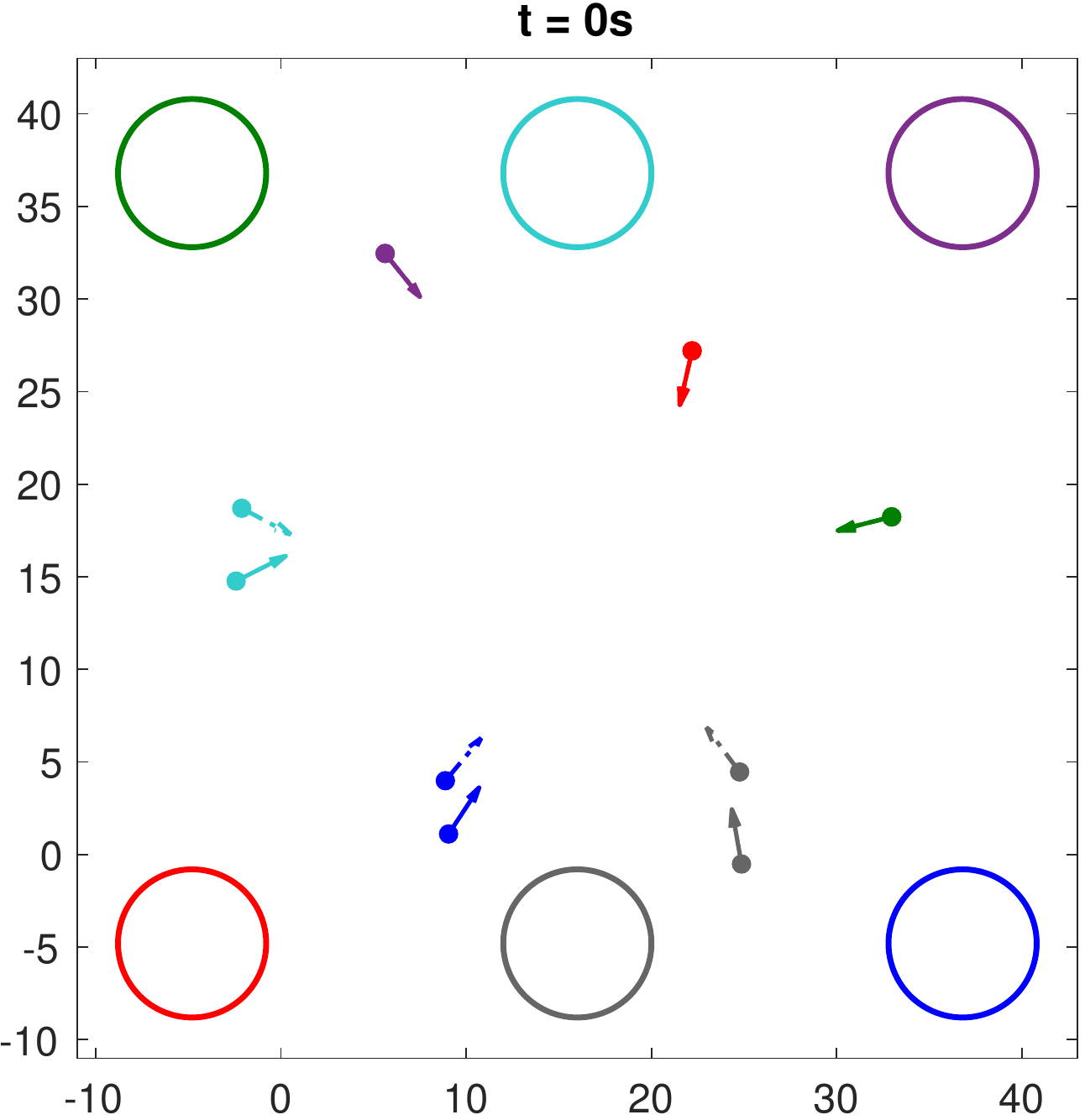}
  \end{subfigure}
  \begin{subfigure}[b]{0.45\textwidth}
    \includegraphics[width=\textwidth]{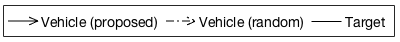}
  \end{subfigure}
  \caption{In this figure, we illustrate the initial states selected by our proposed learning-based strategy (solid arrows) versus those chosen with the baseline random selection method (dash-dot arrows) for four different scenarios where our proposed method succeeded in getting all vehicles to their goals successfully without any safety violation while the baseline method resulted in safety violations even though the initial states selected from the two methods are very close to each other.}
  \label{fig:init_compare_plots}
\end{figure}

\begin{figure}[]
\centering
  \begin{subfigure}[b]{0.23\textwidth}
    \includegraphics[width=\textwidth]{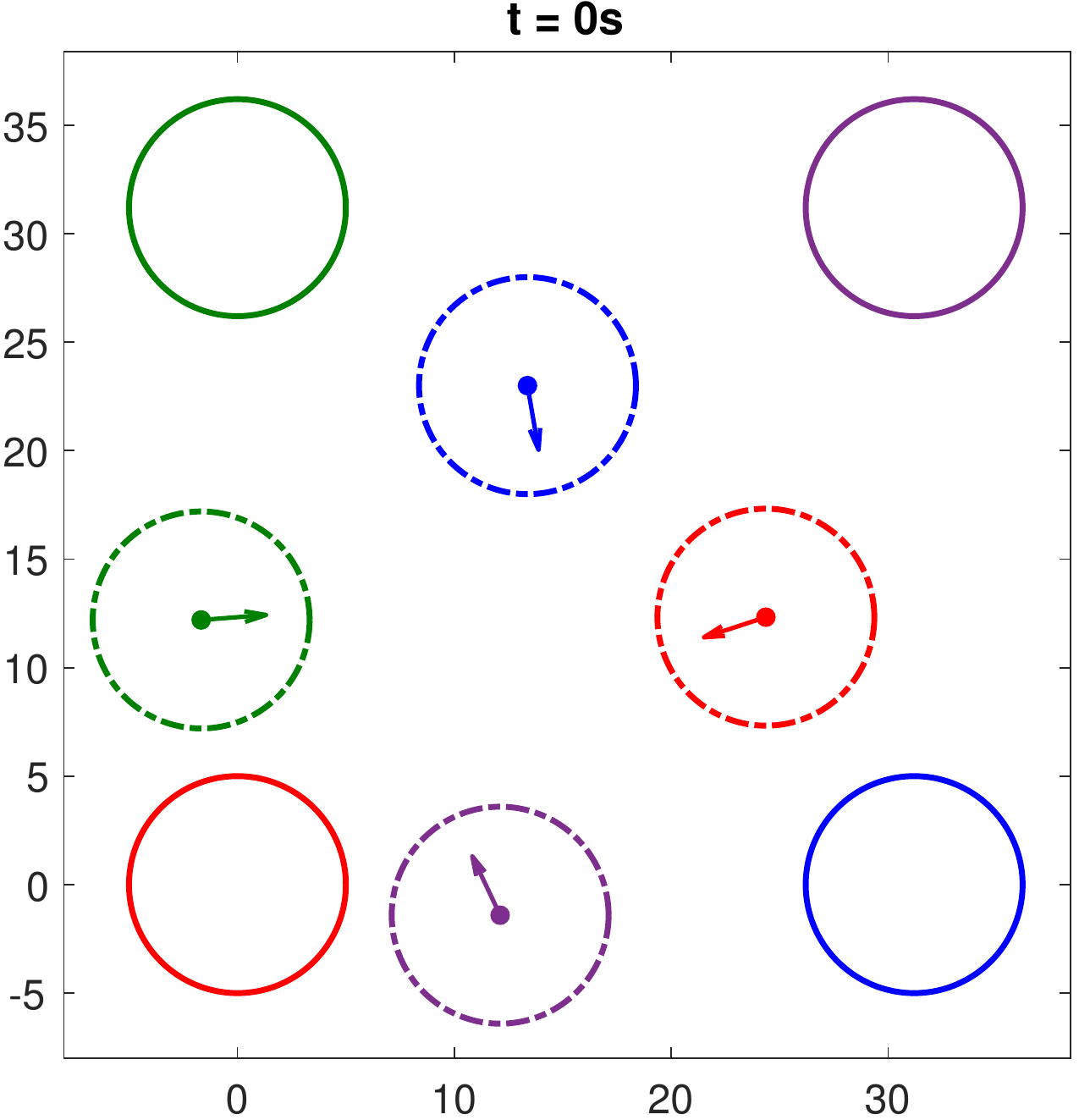}
  \end{subfigure}
  \begin{subfigure}[b]{0.23\textwidth}
    \includegraphics[width=\textwidth]{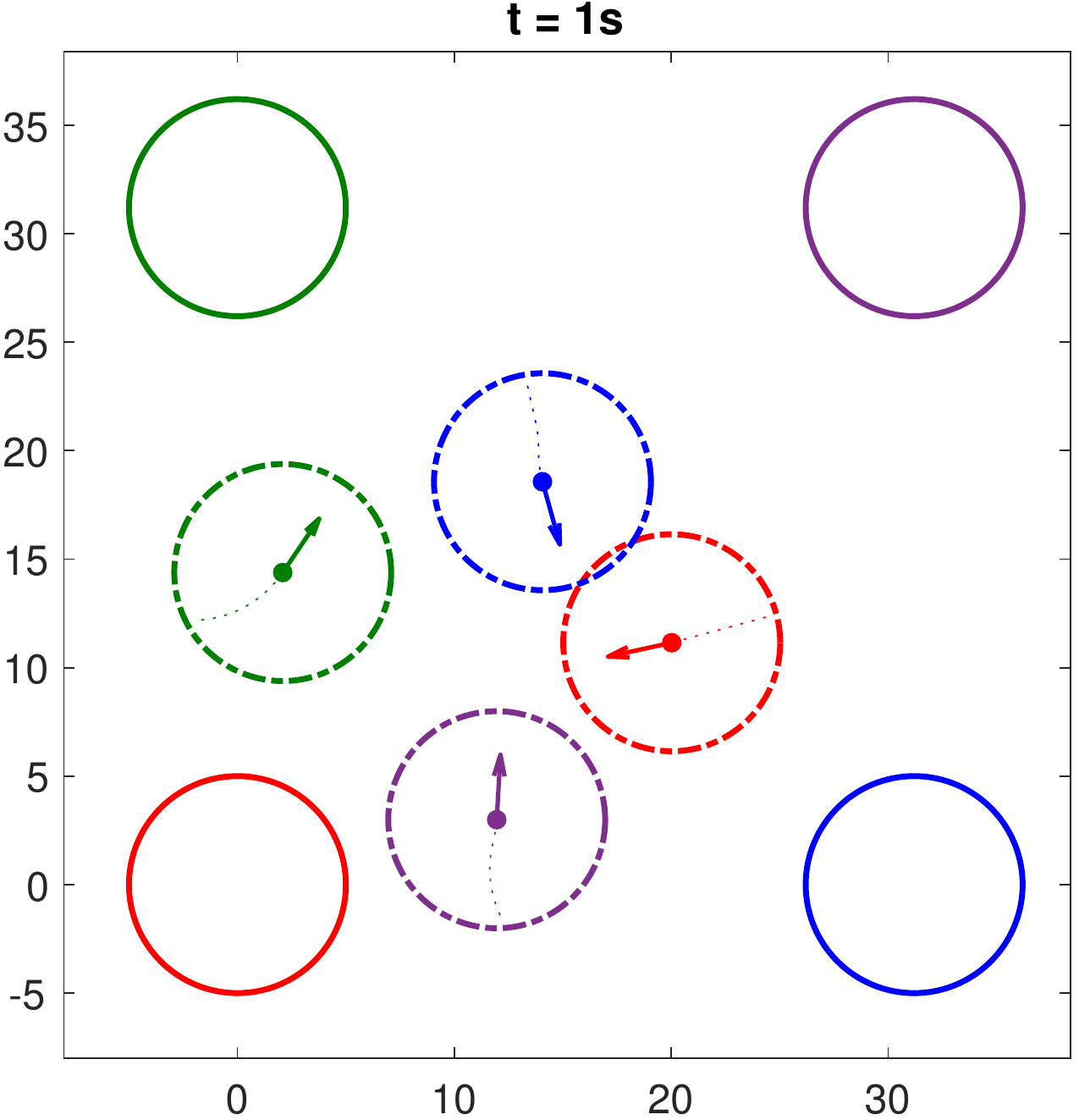}
  \end{subfigure}
  \\
  \begin{subfigure}[b]{0.23\textwidth}
    \includegraphics[width=\textwidth]{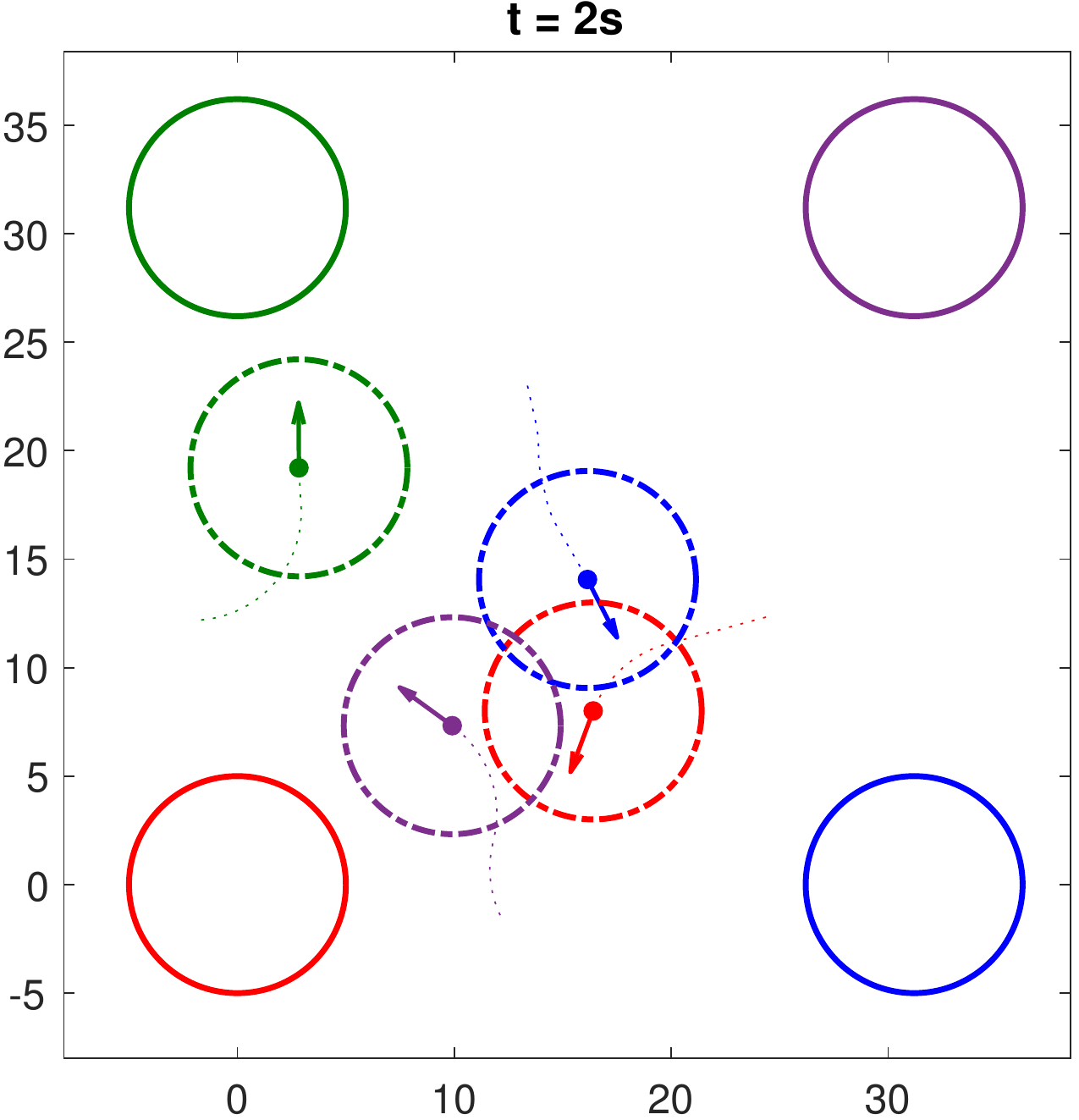}
  \end{subfigure}
  \begin{subfigure}[b]{0.23\textwidth}
    \includegraphics[width=\textwidth]{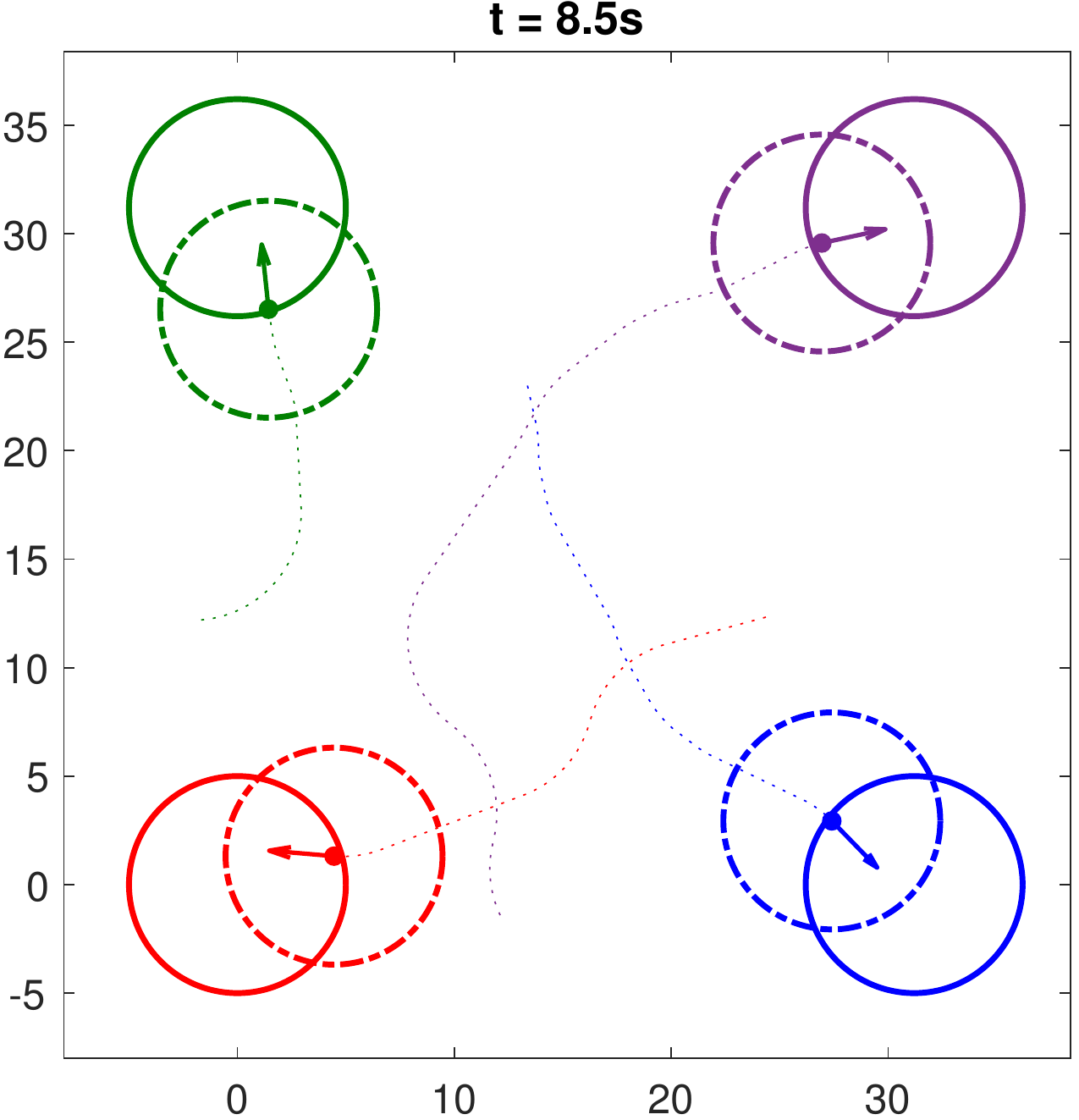}
  \end{subfigure}
  \begin{subfigure}[b]{0.45\textwidth}
    \centering
    \includegraphics[width=\textwidth]{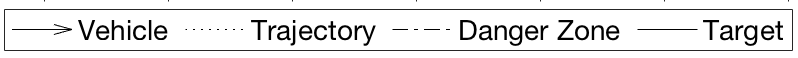}
  \end{subfigure}
  \caption{This figure illustrates different time points of the simulation when we use our proposed learning-based initialization strategy to select the initial states of the vehicles. This scenario is identical to that in the top right figure of Figure \ref{fig:init_compare_plots}. We observe that our proposed approach learns to identify the strength of the safety-aware algorithm in guaranteeing safety for three vehicles and initializes vehicles such that only three vehicles end up coming into close contact with each other. All vehicles successfully reach their goals without \textit{any} safety violations.}
  \label{fig:four_veh_optimized}
\end{figure}

\begin{figure}[]
\centering
  \begin{subfigure}[b]{0.23\textwidth}
    \includegraphics[width=\textwidth]{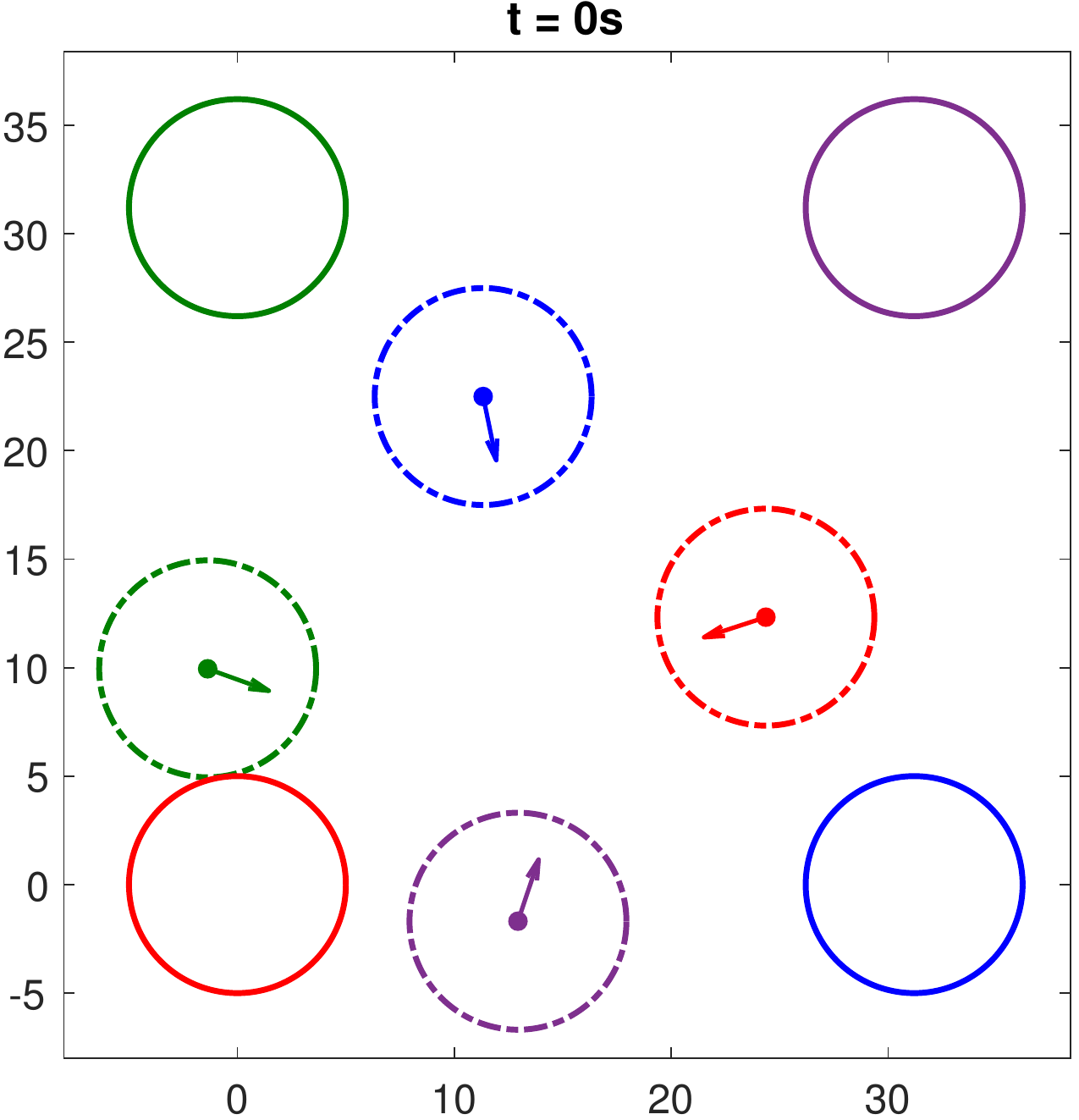}
  \end{subfigure}
  \begin{subfigure}[b]{0.23\textwidth}
    \includegraphics[width=\textwidth]{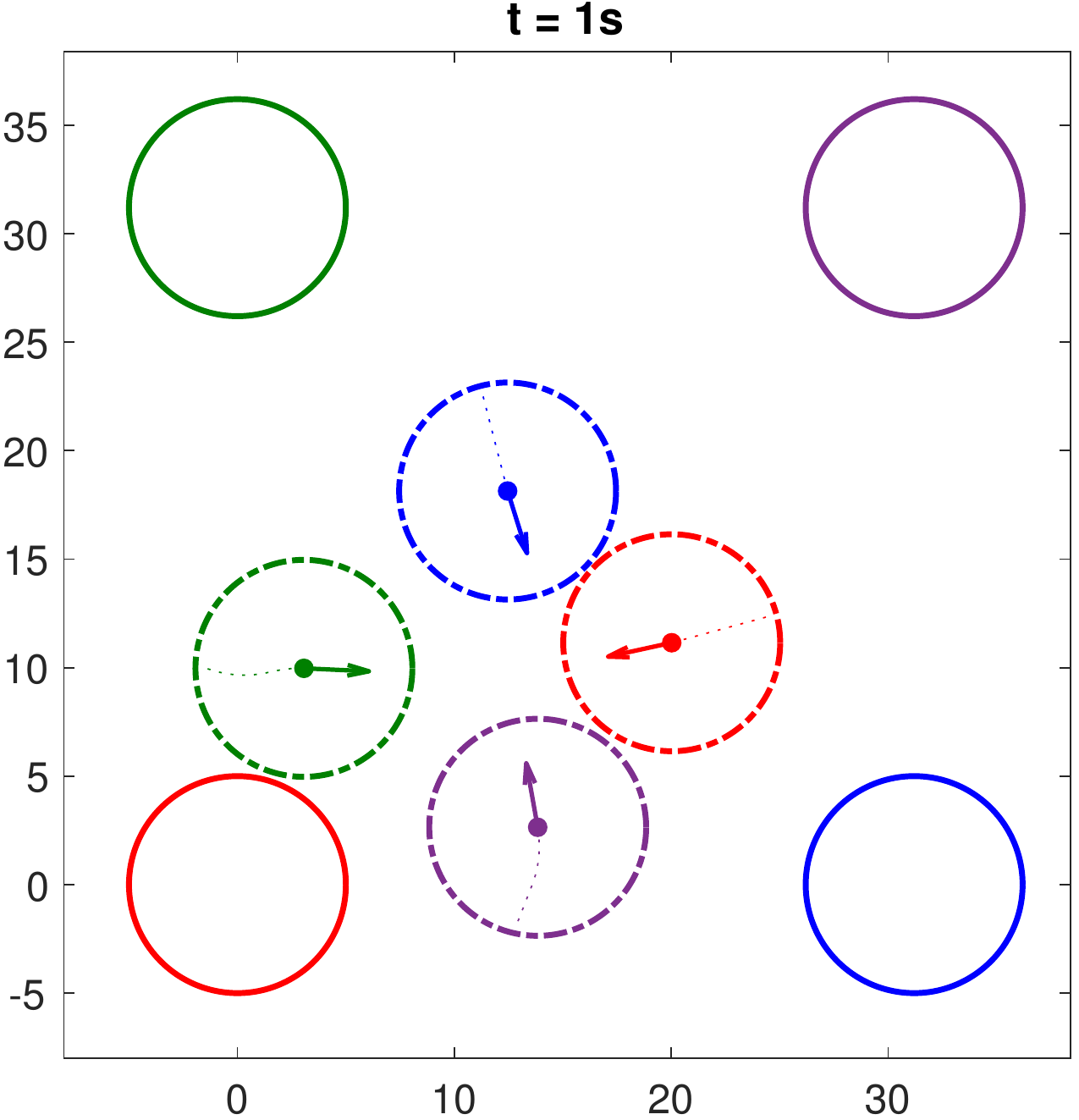}
  \end{subfigure}
  \\
\begin{subfigure}[b]{0.23\textwidth}
    \includegraphics[width=\textwidth]{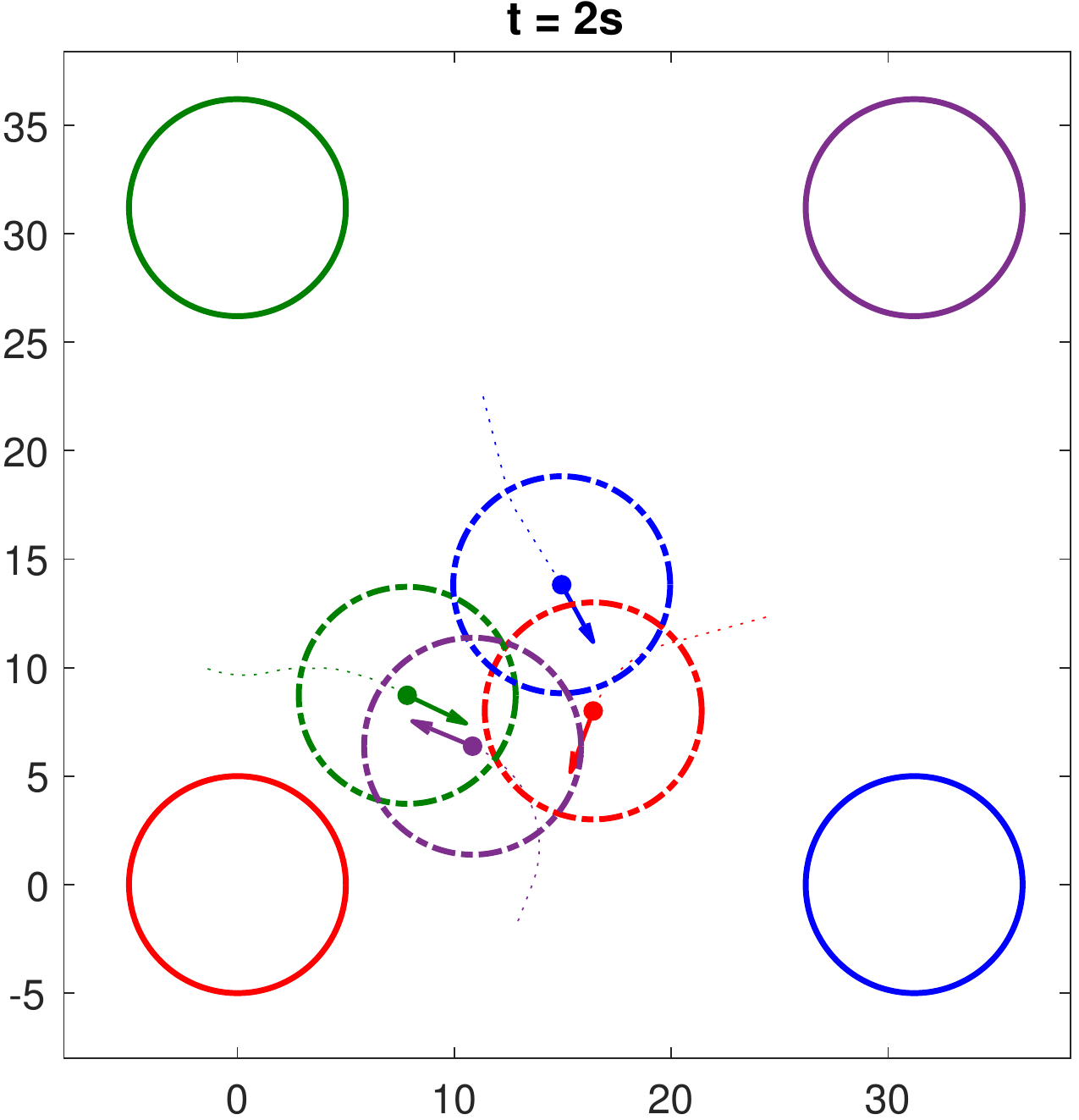}
  \end{subfigure}
  \begin{subfigure}[b]{0.23\textwidth}
    \includegraphics[width=\textwidth]{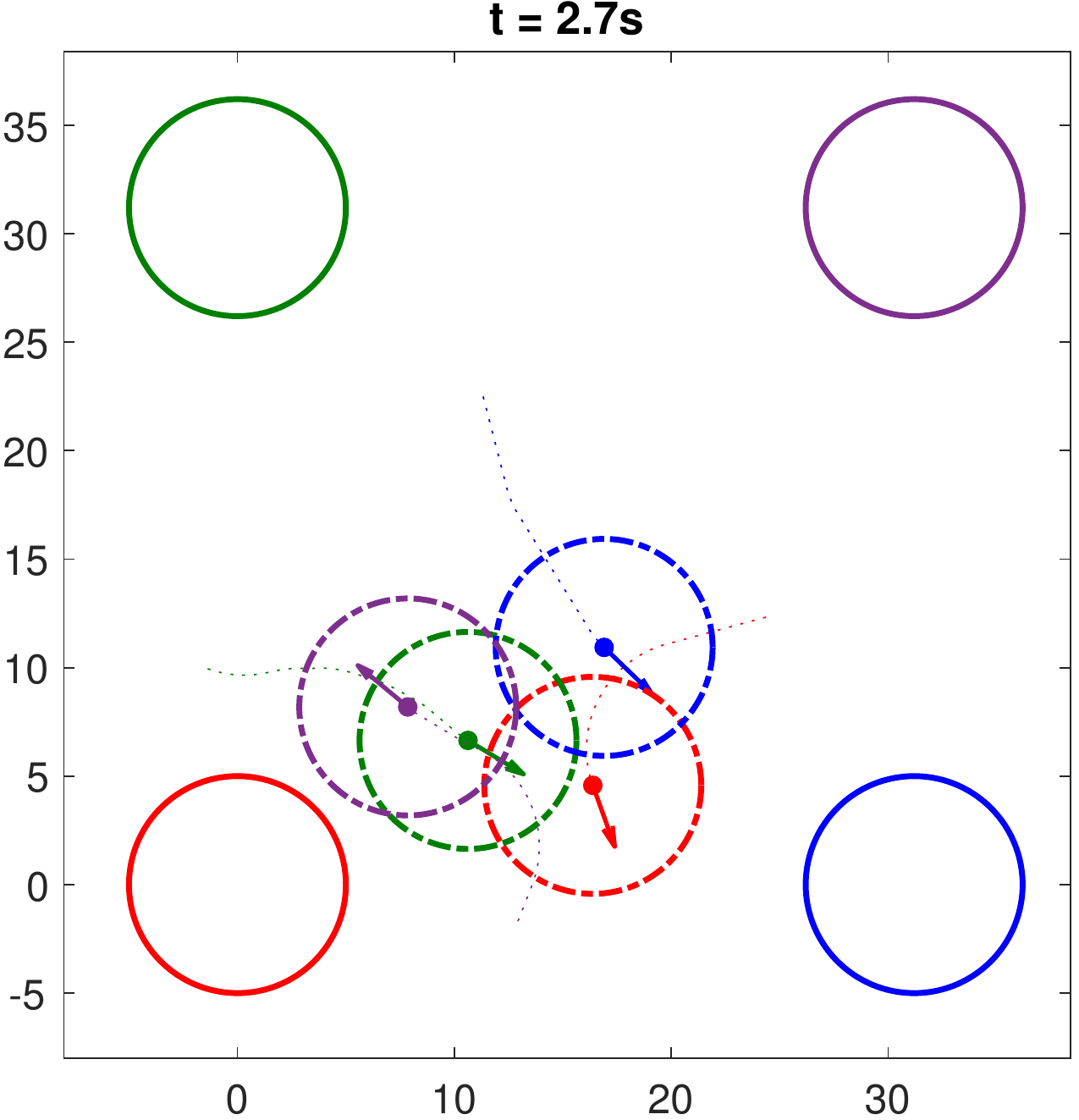}
  \end{subfigure}
  \begin{subfigure}[b]{0.45\textwidth}
    \centering
    \includegraphics[width=\textwidth]{other_plots/four_veh_legend.png}
  \end{subfigure}
  \caption{This figure illustrates the simulation when the baseline random initialization strategy is used in the scenario identical to that in the top right figure of Figure \ref{fig:init_compare_plots}. This is meant to contrast Figure \ref{fig:four_veh_optimized} that with the randomized strategy, even though the initial states are very close to those selected by our proposed method, it results in danger zone violations between the green and the purple vehicles at time $t=2.0$s and $t=2.7s$.}
  \label{fig:four_veh_random}
\end{figure}

In Figure \ref{fig:init_compare_plots}, we plotted the initial states selected with our proposed learning-based method (solid arrows) versus those selected via the baseline randomized selection (dash-dot arrows) from the $\numsample$ candidate sets of initial states sampled around the original proposed initial states for each scenario. For all depicted scenarios, the least-restrictive safety-aware algorithm was able to get all agents successfully to the goal locations without \textit{any} safety violation with our proposed learning-based initialization strategy while the randomized selection failed and resulted in safety violations. The goal location of each vehicle is plotted with the same color as the vehicle. When the initial state of a vehicle is not allowed to be modified, initial states are identical for both strategies and the solid and dash-dot arrows are overlaid on top of each other. 

Although it is not possible for humans to always pinpoint why the initial states learned by our proposed method are more effective for safety than the baseline just by looking at the initial states given that they are generally very close to each other, we can observe patterns by running the least-restrictive safety-aware algorithm. Our approach tends to effectively identify initial states such that vehicles are less likely to run into the situation where many vehicles are on the boundary of the unsafe sets of each other simultaneously or the situation where conflicts of multiple vehicles are less likely to be resolved based on the algorithm used. This shows that our proposed method is able to reason about safety based on the geometry of initial states of the vehicles and their goals by identifying the strength of the safety-aware algorithm used.

Figure \ref{fig:four_veh_optimized} and Figure \ref{fig:four_veh_random} further illustrate the proposed learned v.s. baseline initialization in the scenario depicted in the top right figure of Figure \ref{fig:init_compare_plots}. We plot the danger zone around each vehicle with a dash-dot circle in the same color as the vehicle; if the base of the arrow representing a vehicle is in a circle of a different color, safety has been violated. We see that in Figure \ref{fig:four_veh_optimized}, our proposed initialization strategy enables that only three (red, blue, purple) of the four vehicles get close to the unsafe sets of each other, which our safety-aware algorithm is able to resolve with guaranteed success. At the end, all vehicles get to their goals without \textit{any} safety violations. On the other hand, in Figure \ref{fig:four_veh_random}, the initial states selected by the baseline strategy results in all four vehicles getting very close to each other and the algorithm isn't able to maintain safety while resolving the conflicts. We can see that the green and purple vehicles violated safety at time $t=2.0$s and $t=2.7$s. We see that our learning-based proposed method is able to effectively pick up on the advantages of the least-restrictive safety-aware algorithm  assign a higher probability of success for initialization that is favorable with the algorithm used, effectively improving the safety performance of the multi-vehicle system.

\begin{table*}[h]
\centering
\begin{tabular}{@{}rrrrcrrcrrcrrcrrcrr@{}}\toprule
& & \multicolumn{2}{c}{$\numfixed=0$} && \multicolumn{2}{c}{$\numfixed=1$}  && \multicolumn{2}{c}{$\numfixed=2$}
&& \multicolumn{2}{c}{$\numfixed=3$} && \multicolumn{2}{c}{$\numfixed=4$} && \multicolumn{2}{c}{$\numfixed=5$} \\
\cmidrule{3-4} \cmidrule{6-7} \cmidrule{9-10} \cmidrule{12-13} \cmidrule{15-16} \cmidrule{18-19}
& & $p_{s}$ & $N_{col}$ && $p_{s}$ & $N_{col}$ && $p_{s}$ & $N_{col}$ && $p_{s}$ & $N_{col}$ && 
$p_{s}$ & $N_{col}$ && $p_{s}$ & $N_{col}$
\\ 
\midrule
\multirow{2}{*}{$N=4$} & Learned & \textbf{92} & \textbf{0.425}  && \textbf{88.5} & \textbf{0.915} && \textbf{89.5} & \textbf{0.59} && \textbf{81.5} & \textbf{1.175} && - & - && - & - \\
& Random & 75.5 & 1.58  && 77.5 & 1.69 && 80 &  1.255 && 76 & 1.75 && - & - && - & - \\
\\
\multirow{2}{*}{$N=5$} & Learned & \textbf{88} & \textbf{1.11}  && \textbf{79.5} & \textbf{1.87} && \textbf{80.5} & \textbf{1.44} && \textbf{79.5} & \textbf{1.925} && \textbf{69.5} & \textbf{2.61} && - & - \\
& Random & 66.5 &  3.32 && 64.5 & 3.415 && 67.0 & 2.925 && 73.5 & 2.265 && 57.5 & 4.35 && - & - \\
\\
\multirow{2}{*}{$N=6$} & Learned & \textbf{74} & \textbf{2.54}  && \textbf{67.5} & \textbf{3.145} && \textbf{70.5} & \textbf{2.46} && \textbf{66.5} & \textbf{3.5} && \textbf{63} & \textbf{3.635} && \textbf{62} & \textbf{3.575} \\
& Random & 66 & 3.77  && 56 & 4.58 && 62 & 4.335 && 61.5 & 4.285 && 55 & 4.575 && 58 & 4.425 \\
\\
\bottomrule
\end{tabular}
\caption{\small In this table, we summarize the success rate $p_s$ and the average number of collisions $N_{col}$ where speed $v = 6$ and danger zone radius $R_c=4$ when using the learned initialization strategy versus using the baseline randomized initialization strategy for number of vehicles $N=4, 5, 6$. Our method outperforms the baseline in safety performance for both metrics across all scenarios.}
\label{tab:speed_6_r_4}
\end{table*}

\begin{table*}[h]
\centering
\begin{tabular}{@{}rrrrcrrcrrcrrcrrcrr@{}}\toprule
& & \multicolumn{2}{c}{$\numfixed=0$} && \multicolumn{2}{c}{$\numfixed=1$}  && \multicolumn{2}{c}{$\numfixed=2$}
&& \multicolumn{2}{c}{$\numfixed=3$} && \multicolumn{2}{c}{$\numfixed=4$} && \multicolumn{2}{c}{$\numfixed=5$} \\
\cmidrule{3-4} \cmidrule{6-7} \cmidrule{9-10} \cmidrule{12-13} \cmidrule{15-16} \cmidrule{18-19}
& & $p_{s}$ & $N_{col}$ && $p_{s}$ & $N_{col}$ && $p_{s}$ & $N_{col}$ && $p_{s}$ & $N_{col}$ && 
$p_{s}$ & $N_{col}$ && $p_{s}$ & $N_{col}$
\\ 
\midrule
\multirow{2}{*}{$N=4$} & Learned & \textbf{91} &  \textbf{0.785}  && \textbf{90} & \textbf{0.875} && \textbf{80.5} & \textbf{2.715} && \textbf{80} & \textbf{1.68} && - & - && - & - \\
& Random & 68 & 3.305  && 66.5 & 3.56 && 73 &  2.88 && 71 & 2.81 && - & - && - & - \\
\\
\multirow{2}{*}{$N=5$} & Learned & \textbf{80} & \textbf{2.315}  && \textbf{77.5} & \textbf{2.79} && \textbf{71} & \textbf{4.01} && \textbf{74} & \textbf{3.39} && \textbf{71} & \textbf{3.51} && - & - \\
& Random & 65 &  4.55 && 62 & 5.39 && 53.5 & 8.2 && 58.5 & 5.77 && 61 & 4.94 && - & - \\
\\
\multirow{2}{*}{$N=6$} & Learned & \textbf{65.5} & \textbf{4.065}  && \textbf{64} & \textbf{6.375} && \textbf{61.5} & \textbf{5.36} && \textbf{64} & \textbf{5.11} && \textbf{59} & \textbf{5.805} && \textbf{57.5} & \textbf{6.31} \\
& Random & 57.5 & 6.075  && 46.5 & 8.69 && 52 & 8.27 && 52 & 7.30 && 49 & 7.25 && 53.5 & 7.70 \\
\\
\bottomrule
\end{tabular}
\caption{\small This table summarizes the success rate $p_s$ and the average number of collisions $N_{col}$ where speed $v = 5$ and danger zone radius $R_c=5$ for number of vehicles $N=4, 5, 6$. Similarly, we see that our proposed learning-based strategy outperforms the baseline in terms of safety performance across all scenarios.}
\label{tab:speed_5_r_5}
\vspace{-2.5em}
\end{table*}

During the training phase, all models take less than $30$ seconds to complete training. During the online phase where we figure out the new initial states based on the learned model and the original proposed initial states, it takes on average $0.30$ seconds total to compute the optimal sets of initial states for all $\numtrials=200$ runs in parallel for a given $(v, R_c, \numvehicle, \numfixed)$ setting with our proposed strategy. Thus our proposed method can be very efficiently applied as we encounter new scenes online.

\section{Conclusion}
In this paper, we proposed a novel approach for enhancing the safety performance of least-restrictive safety-aware algorithms for multiple vehicles in \textit{unstructured} settings and showed that it is possible to use machine learning to make \textit{minor} modifications to initial states of vehicles in the environment and improve,  sometimes quite substantially, the safety of the system compared to a randomized initialization approach. This is a promising step towards making least-restrictive safety algorithms such as those enabled by reachability more practically useful in unstructured scenarios for multi-vehicle systems that safety cannot be guaranteed for with the algorithms. 

 \bibliographystyle{IEEEtran}
 \bibliography{references}
  
 \clearpage

\end{document}